# Limit-behavior of a hybrid evolutionary algorithm for the Hasofer-Lind reliability index problem

Gonçalo das Neves Carneiro
INEGI / LAETA
*Faculty of Engineering, University of Porto*
Porto, Portugal
gncarneiro@fe.up.pt

Carlos Conceição António
INEGI / LAETA
*Faculty of Engineering, University of Porto*
Porto, Portugal
cantonio@fe.up.pt

*Abstract—* In probabilistic structural mechanics, the Hasofer-Lind reliability index problem is a paradigmatic equality constrained problem of searching for the minimum distance from a point to a surface. In practical engineering problems, such surface is defined implicitly, requiring the solution of a boundary-value problem. Recently, it was proposed in the literature a hybrid micro-genetic algorithm (HmGA), with mixed real-binary genotype and novel deterministic operators for equality-constraint handling, namely the Genetic Repair and Region Zooming mechanisms (G. das Neves Carneiro and C. Conceição António, "Global optimal reliability index of implicit composite laminate structures by evolutionary algorithms", Struct Saf, vol. 79, pp. 54-65, 2019). We investigate the limit-behavior of the HmGA and present the convergence theorems for the algorithm. It is proven that Genetic Repair is a conditionally stable mechanism, and its modes of convergence are discussed. Based on a Markov chain analysis, the conditions for the convergence with probability 1 of the HmGA are given and discussed.

## I. Introduction

In structural engineering, the existence of uncertainty in the main design variables and material properties is frequently taken into consideration. Monte Carlo Simulation (MCS) is the foundation for sampling methods in probabilistic structural reliability. It consists in evaluating a traditional binary process, for which the frequency of failure events is an unbiased estimator of the probability of failure [1]. In practical design problems, MCS may be prohibitive due to very small probabilities and the need to evaluate implicit limit-state functions.

Alternatively, the Hasofer-Lind reliability index (HL-RI) problem allows to compute approximate values of the probability of failure, via the solution of an equality constrained minimization problem [2,3]. Most methods to estimate the reliability index are in the form of gradient-based algorithms [2-7]. Despite their efficiency, gradient-based methods often diverge for high dimensional problems, highly concave limit-state functions and/or non-smooth implicit response functionals [7-11].

The use of Evolutionary Algorithms (EAs) to find the global reliability index in structural problems is still scarce, possibly due to the natural inability of EAs to handle equality constraints. Cheng and Li [12] proposed an artificial neural network (ANN)-based genetic algorithm (GA). The method involves the generation of data sets by the Uniform Design Method to train an ANN as a surrogate of the limit-state function. Results are promising but limited to low dimensional and analytical examples. Shao and Murotsu [13] proposed a selective search strategy of failure modes, limited to a set of search directions at 45º angle intervals, and used the simple GA to find the critical reliability index among the possible combination of failure modes. Deng et al. [14] extended the previous work by developing a shredding GA, applying a filtration process of solutions with undesired characteristics. Wang and Ghosn [15] further modified the previous shredding GA to improve convergence, based on a linkage-learning process.

Recently, das Neves Carneiro and Conceição António [16-18] proposed a hybrid micro-Genetic Algorithm (HmGA) to solve the HL-RI problem, with specific deterministic operators for equality-constraint handling, namely the Genetic Repair and the Region Zooming mechanisms. The algorithm further relies on the dynamics between highly elitist and highly disruptive stochastic operators and works on a mixed real-binary genotype space. The HmGA was validated against 42 designs of composite laminate shell structures with known reliability index.

In this article, we investigate the limit-behavior of the HmGA [16-18] and present the convergence theorems for the algorithm, giving a rigorous and unambiguous proof of its ability to converge. It is demonstrated that the Genetic Repair mechanism is a conditionally stable iterative process and based on a Markov Chain analysis the conditions for the convergence with probability 1 of the HmGA are given.

The article is organized as follows. In Sect. II the HL-RI problem is succinctly introduced. In Sect. III the equality constraint handling mechanisms are formally defined and analyzed. In Sect IV a set of diverse stochastic evolutionary operators is described. In Sect. V the limit-behavior of the HmGA is analyzed.

## II. The Hasofer-Lind Reliability Index

In structural design, *reliability* is commonly interpreted as a measure of the proximity to failure of structural systems. Conceptually, reliability is closely related with the notion of feasibility in the design space and failure is understood as the state of a system violating the integrity constraints (limit-states) imposed in the design problem.

Let $U \subseteq \mathbb{R}^M$ be the state space of state variables $\mathbf{u}$ and $\Omega \subseteq \mathbb{R}^N$ a sample space of random variables $\mathbf{x}$. Structural equilibrium is defined by a mapping $\Psi: U \times \Omega \to U$, such that an (implicit) state equation:

$$\Psi(\mathbf{u}(\mathbf{x}), \mathbf{x}) = 0 \qquad (1)$$

holds. A functional $\gamma: U \times \Omega \to \mathbb{R}$, with both implicit and explicit dependence on $\mathbf{x}$, i.e., $g(\mathbf{x}) = \gamma(\mathbf{u}(\mathbf{x}), \mathbf{x})$, is called *limit-state function* and gives a real-valued measure of the system's state. The continuity of $g$ is assumed throughout the manuscript.

In structural reliability, a probability space $(\Omega, \mathcal{F}, P)$, with $\mathcal{F} = \{\emptyset, D_f, D_s, \Omega\}$, is called an *uncertainty space*, where $D_f = \{\mathbf{x} \in \Omega : g(\mathbf{x}) < 0\}$ and $D_s = \overline{D_f}$ are disjoint subsets of $\Omega$, called the *failure space* and the *safety space*, respectively.

The structural *probability of failure* is thus defined as the probability of a design solution being on the failure space, due to uncertainty in the design variables/parameters, i.e. [1]:

$$p_f = \Pr(g(\mathbf{x}) < 0) = \int_{g(\mathbf{x})<0} p_\mathbf{x}(\mathbf{x})\, d\mathbf{x} \quad (2)$$

where $p_\mathbf{x}(\mathbf{x})$ is the joint probability density function (PDF) of the random variables. In real-scenario structural problems, equation (1) is solved numerically, usually by the Finite Elements Method. Thus, the integral in (2) can only be solved numerically as well.

Numerical reliability assessment methods are divided into *global* and *local* methods. *Global reliability methods* analyze the whole sample space $\Omega$, or a relevant part of it, to provide an unbiased estimator of the probability of failure. MCS is the most straightforward approach and sets the basis for other sampling methods. *Local reliability methods* focus their action on the vicinity of the mean-values of the random variables. These methods obviate the integration process by transforming $p_\mathbf{x}(\mathbf{x})$ into a multi-normal joint PDF, with known properties [1].

In structural design optimization, the repeated application of global methods to compute extremely low probabilities of failure requires the exhaustive evaluation of the sample space. In the context of structural design optimization, global methods often become impractical. For that reason, local methods are still relevant and applied in engineering design problems.

Hasofer and Lind [2] proposed an invariant reliability measure defined at the point of the failure surface with the greatest probability density. That is, the point around which failure events are most likely to occur. In the space of *independent standard normal random variables* $\mathbf{y} \sim N(\mathbf{0}, \mathbf{1})$, the identification of such point is strictly a minimization problem, given by:

$$\min_{\mathbf{y}} \ \|\mathbf{y}\| \quad (3)$$
$$\text{subject to:} \quad G(\mathbf{y}) = 0$$

where $\mathbf{y} = T(\mathbf{x})$, with $T: \Omega \to Y$ and $Y \subseteq \mathbb{R}^N$, is an invertible transformation and $G(\mathbf{y}) = g(T^{-1}(\mathbf{y}))$. The solution to problem (3) is called the *most probable failure point* (MPP), $\mathbf{y}_{\text{MPP}}$, and:

$$\beta_{HL} = \|\mathbf{y}_{\text{MPP}}\| \quad (4)$$

is the *Hasofer-Lind reliability index* (HL-RI), representing the shortest distance, in standard-deviation units, from the origin of the standardized uncertainty space to the failure surface. A geometric representation of the reliability index problem can be seen in Fig. 1.

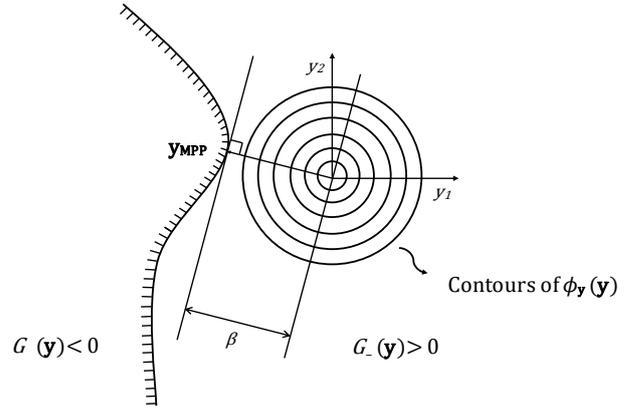

Fig. 1. Two-dimensional representation of the Hasofer-Lind problem, where $\phi_\mathbf{y}$ is the standard multinormal probability density function.

Given the optimality of $\mathbf{y}_{\text{MPP}}$, the reliability index $\beta_{HL}$ is unique [2] and hence invariant. It then results that, by approximating the failure surface as a hyperplane, at $\mathbf{y}_{\text{MPP}}$, the probability of failure is approximately equal to [1]:

$$p_f \simeq \Phi(-\beta_{HL}) \quad (5)$$

where $\Phi$ is the standard normal cumulative distribution function.

### III. EQUALITY CONSTRAINT HANDLING

Due to an implicit state equation in (1), the HL-RI problem is quite challenging to solve in real-scenario structural reliability problems, asking for a compromise between accuracy and efficiency.

One major difficulty is the lack of knowledge about the failure surface. In structural problems, the failure surface is commonly defined as an ellipsoid in the three-dimensional stress space by means of a quadratic equation. It exists in all directions. However, in the uncertainty space, the failure surface may have different shapes, depending on the functional relationship between the random variables and the stress components. Thus, an additional secondary objective for the numerical solution of the HL-RI problem is posed: *to identify the directions for which the failure surface exists.*

Through the following subsections, a detailed description of the equality constraint handling mechanisms is presented, First, a mixed genotype data structure is defined, followed by a penalty formulation of the HL-RI problem. Then, the Genetic Repair mechanism is analyzed, and conditions are given for its numerical stability and convergence. Finally, the Region Zooming mechanism is discussed.

#### A. Data Structure

An important step in evolutionary computation is the definition of an appropriate data structure to represent the search variables and further manipulate the solutions, during the search process. Consider the following general definitions.

Let $\mathbb{S}$ be a finite search space with any kind of data structure (genotype space) and $\mathbb{A}$ a potentially empty set containing additional search information. A genotype (or solution) is an element $s \in \mathbb{A} \times \mathbb{S}$.

Given that the HL-RI problem can be understood as the problem of finding both the direction and the distance from a point to a surface, the search variables are decomposed into their magnitude and direction components, as follows:

$$\mathbf{y} = \beta \mathbf{a} \quad (6)$$

where $\beta = \|\mathbf{y}\|$ and $\mathbf{a} \in [-1,1]^N$ with $\|\mathbf{a}\| = 1$. Also, please consider the notation $g(\beta, \mathbf{a})$ to represent $g(T^{-1}(\mathbf{y}))$.

From (6), the concept of a *mixed genotype* data structure is introduced to represent points in the standardized uncertainty space and allowing for the manipulation of $\beta$ and the elements of $\mathbf{a}$ as separate search variables of the problem.

**Definition 1:** Let $\mathbb{B}^l$ be the set of binary strings of length $l$, whose elements $b(\mathbf{a})$ are the binary form of the elements in unit vector $\mathbf{a}$. An array $s = (\beta, b(\mathbf{a})) \in \mathbb{R}_{\geq 0} \times \mathbb{B}^l$ is called a *mixed genotype*.

*B. Penalty Formulation*

Order withing the elements of a population is established by means of a real-valued positive fitness measure. A proper definition of fitness is problem-dependent, but important for the overall efficiency of an EA. The primal problem, in (3), is here formulated as a penalty problem for the unconstrained maximization of a fitness function $f: \mathbb{R}_{\geq 0} \times \mathbb{B}^l \to \mathbb{R}$ as follows:

$$\max_{\beta, \mathbf{a}} \; f = C - \beta - \lambda \Gamma\left(g(T^{-1}(\mathbf{y}))\right) \quad (7)$$

where $C$ is a positive constant, $\lambda$ is a scaling factor and $\Gamma(\cdot)$ is a non-negative penalty function, given by:

$$\Gamma(g) = \begin{cases} K|g|^q & , \text{if } |g| > \eta \\ 0 & , \text{otherwise} \end{cases} \quad (8)$$

where $\eta > 0$ is a small tolerance and both $K, q > 0$ are determined considering two degrees of violation, as in [19].

From the natural order of the real numbers, it follows.

**Definition 2:** Let $\mathbb{S}^*$ be the set of all finite lists over $\mathbb{R}_{\geq 0} \times \mathbb{B}^l$. A population $\mathbf{P}^t \in \mathbb{S}^*$, at time $t$, is a list $(s_1, s_2, \ldots, s_{n_\mathbf{P}}) \in \mathbb{S}^*$ of size $n_\mathbf{P}$, ordered in terms of fitness and allowing repetition. A list $\mathbf{E}^t \subset \mathbf{P}^t$, containing the mixed genotypes with the highest fitness, defines the elite group of the population.

*C. Genetic Repair Mechanism*

To repair a genotype with an infeasible phenotype means to transform that same genotype into another with a feasible phenotype. The repair may be *total* if the constraint violation is eliminated, or *partial*, if reduced.

The Genetic Repair mechanism (GRM) acts on each solution by manipulating the real-valued gene $\beta$, while the binary-valued genes $b(\mathbf{a})$ are kept frozen. It is a Lamarckian-kind operator, promoting the local improvement of infeasible solutions. It also allows the identification of good search directions, as those for which a total genetic repair is possible.

Given an infeasible solution, the mechanism consists of a deterministic iterative process that reduces the limit-state function until zero, along a fixed direction.

Let $(\beta^k)_{k \geq 0}$ be a sequence of partial sums generated by the following recurrence relation:

$$\beta^{k+1} = \begin{cases} \beta^k + \Delta \beta^k & , \text{if } g(\beta^k|\mathbf{a}) \geq 0 \\ \beta^k - \Delta \beta^k & , \text{otherwise} \end{cases} \quad (9)$$

where:

$$\Delta \beta^k = \Delta_{\max}^k \left( 2\, e^{\frac{|g(\beta^k|\mathbf{a})| - g_0}{g_0} \ln(2)} - 1 \right) \quad (10)$$

is a nonlinear increment function, with $g_0 = g(T^{-1}(\mathbf{0}))$, $|g(\beta^k|\mathbf{a})|$ is the absolute value of $g$, calculated at $\beta^k$, given a unit direction $\mathbf{a}$, and $\Delta_{\max}^k > 0$ is an amplitude parameter.

From $(\beta^k)_{k \geq 0}$, a naturally occurring sequence $(g^k)_{k \geq 0}$, with $g^k = |g(\beta^k|\mathbf{a})|$, is generated. Hence, the GRM is said to converge if $\lim_{k \to \infty} g^k = 0$ and $\lim_{k \to \infty} \beta^k = \bar{\beta}$.

We begin by analyzing sequence $(g^k)_{k \geq 0}$. A sufficient condition for its convergence is that the sequence be both bounded and strictly decreasing (*monotone convergence theorem*). For that, it suffices to show that:

$$0 < |g(\beta^k|\mathbf{a})| \leq M \quad (11)$$

and

$$|g(\beta^{k+1}|\mathbf{a})| < |g(\beta^k|\mathbf{a})| \quad (12)$$

for all $k \geq 0$ and for some $M < \infty$.

Regarding the stability of the method, notice that for a given value of $\Delta_{\max}^k$, the increment function may not guarantee the strict monotonicity of $(g^k)_{k \geq 0}$, at every two consecutive iterations. Indeed, it is the parameter $\Delta_{\max}^k$ that controls both the efficiency and the numerical stability of the GRM. If too small, the iterative process becomes cumbersome. If too large, stability (strict monotonicity) may be lost, and the sequence diverge.

A necessary and sufficient condition for stability follows.

**Proposition 1:** For all $k \geq 0$, $|g(\beta^{k+1}|\mathbf{a})| < |g(\beta^k|\mathbf{a})|$ if and only if $\Delta_{\max}^k$ is bounded.

**Proof:** By definition $\Delta_{\max}^k > 0$. Also, the increment function is nonnegative, hence from (9) it results $\Delta \beta^k = |\beta^{k+1} - \beta^k|$. Now, let $(g^k)_{k \geq 0}$ be strictly decreasing. Solving (10) for $|g(\beta^k|\mathbf{a})|$ and substituting on the right-hand side of (12), it yields:

$$\left( 2 e^{\frac{|g(\beta^{k+1}|\mathbf{a})| - g_0}{g_0} \ln(2)} - 1 \right) \Delta_{\max}^k < |\beta^{k+1} - \beta^k| \quad (13)$$

Since $g$ is a continuous real-valued function, both sides of (13) are finite and so $\Delta_{\max}^k$ is bounded from above, $\forall k \geq 0$.

Conversely, for a nondecreasing sequence, the opposite to inequality (13) is obtained (i.e., with '$\geq$') and so $\Delta_{\max}^k$ is unbounded from above, for all $k \geq 0$. ∎

A few comments on the practical implications of the previous result. A value of $\Delta_{\max}^k$ satisfying inequality (13) is said to be *sufficiently small*. If not, and a strictly decreasing sequence can be generated from $g(\beta|\mathbf{a})$, then it must be adjusted. Moreover, the inequality is nonlinear. As such, it does not define explicit upper bounds on $\Delta_{\max}^k$. Instead, a change in the value of $\Delta_{\max}^k$ implies a change in the values of $\beta^{k+1}$ and $|g(\beta^{k+1}|\mathbf{a})|$. When not satisfied, the inequality must be evaluated iteratively until it holds. The GRM is, therefore, a conditionally stable method.

Inequality (13) refers to iteration $k$. As such, it does not imply that the increment $\Delta\beta^{k+1}$, at iteration $(k+1)$, be necessarily smaller than the increment $\Delta\beta^k$, at iteration $k$. This is true only if the value of $\Delta_{\max}^k$ is kept constant while being sufficiently small through consecutive iterations.

Inequality (13) does not make any consideration about function $g$, other than it allowing to generate strictly decreasing sequences. Indeed, $g(\beta|\mathbf{a})$ need not be monotonic, but the sequence generated from it must. Notice that a sufficiently small $\Delta_{\max}^k$ alone does not guarantee the convergence of sequence $(g^k)_{k\geq 0}$ to zero, since it allows convergence to stationary points of $g(\beta|\mathbf{a})$.

As such, conditions on $g(\beta|\mathbf{a})$ are required. Consider the following.

**Definition 3:** Let $\bar{\beta}$ be the minimum of a function, on a given interval. A function is said *unimodal* if and only if it is strictly decreasing for $\beta \leq \bar{\beta}$ and strictly increasing for $\bar{\beta} \geq \beta$, on the same interval.

A sufficient condition for the convergence of sequence $(g^k)_{k\geq 0}$ is now established.

**Theorem 1:** Let $g(\beta|\mathbf{a})$ be strictly decreasing on some closed interval $[\beta^0, \beta^u]$. If $\exists \bar{\beta} \in (\beta^0, \beta^u)$, such that $g(\bar{\beta}|\mathbf{a}) = 0$, and $\Delta_{\max}^k$ is sufficiently small, for all $k \geq 0$, then sequence $(g^k)_{k\geq 0}$ converges to zero.

**Proof:** The continuity and strict monotonicity of $g(\beta|\mathbf{a})$ imply that $|g(\beta|\mathbf{a})|$ be bounded and unimodal on the same interval, with $\min|g(\bar{\beta}|\mathbf{a})| = 0$, at $\bar{\beta}$.

Thus, if $\Delta_{\max}^k$ is sufficiently small, then (12) holds and there exists an $M < \infty$ such that $|g(\beta^k|\mathbf{a})| < M$, for all $k > 0$. Thus, sequence $(g^k)_{k\geq 0}$ is bounded from above. Moreover, $k \to \infty \Rightarrow |g(\beta^k|\mathbf{a})| \to 0$ and the sequence is bounded from below and will converge to zero. ∎

In theory, decreasing sequences need not be bounded from above to converge but, in this case, guaranteeing the existence of an upper bounded implies that in practice $\Delta_{\max}^k$ need not be close to zero for it to be sufficiently small. That is, between consecutive iterations, the values of $\Delta_{\max}^k$ that are sufficiently small are expected to be 'close' to each other.

**Theorem 2:** Under the same conditions of Theorem 1:

$$\lim_{k\to\infty}|g(\beta^k|\mathbf{a})| = 0 \implies \lim_{k\to\infty}\Delta\beta^k = 0 \quad (14)$$

and hence $\lim_{k\to\infty}\beta^k = \bar{\beta}$.

**Proof:** Given $(g^k)_{k\geq 0}$ convergent to zero, relation (14) follows immediately from (10), implying that the necessary condition for the convergence of the sequence of partial sums $(\beta^k)_{k\geq 0}$ holds. Then, by applying the *ratio test*, it follows that:

$$\lim_{k\to\infty}\left|\frac{\Delta\beta^{k+1}}{\Delta\beta^k}\right| < 1 \quad (15)$$

since the strict monotonicity of $g(\beta|\mathbf{a})$ implies $\Delta_{\max}^{k+1} \leq \Delta_{\max}^k$ and $|g(\beta^{k+1}|\mathbf{a})| < |g(\beta^k|\mathbf{a})|$, for all $k \geq 0$. Thus, sequence $(\beta^k)_{k\geq 0}$ converges absolutely and because $|g(\beta^k|\mathbf{a})|$ converges to zero, then $\beta^k$ must converge to $\bar{\beta}$. ∎

Regarding the behavior of the GRM towards convergence, it is possible to identify two main modes of convergence, both satisfying the conditions of Theorem 1.

*Strong convergence* represents the ideal behavior designed for the mechanism, where it describes a descent and asymptotic path on the positive side of $g(\beta|\mathbf{a})$, starting from an initial estimate $g(\beta^0|\mathbf{a}) = M$ towards $g(\bar{\beta}|\mathbf{a}) = 0$.

*Weak convergence* consists in the oscillatory and asymptotic path, around $g(\bar{\beta}|\mathbf{a}) = 0$, starting from a positive initial estimate $g(\beta^0|\mathbf{a}) = M$ and characterized by sequentially smaller values of $|g(\beta^k|\mathbf{a})|$, for all $k \geq 0$. This oscillatory behavior originates in increments $\Delta\beta^k$ that are large enough to alternate between positive and negative values of $g(\beta|\mathbf{a})$, in consecutive iterations, while $\Delta_{\max}^k$ is still sufficiently small.

Under the conditions of Theorem 1, whenever $\Delta_{\max}^k$ is not sufficiently small, it must be updated. Given the monotonicity of $g(\beta|\mathbf{a})$, then it must be reduced. As a principle, $\Delta\beta^k$ cannot be larger than $\bar{\beta}$. So, a practical rule is to iterate:

$$\Delta_{\max}^k = \alpha\,\beta^k, \quad \alpha \in (0,1) \quad (16)$$

for different values of $\alpha$, until $\Delta_{\max}^k$ is sufficiently small.

Overall, the proposed iterative process runs efficiently towards convergence, varying from coarser to more refined increments, as $g(\beta|\mathbf{a})$ approaches zero. In practice, the mechanism stops when:

$$|g(\beta^k|\mathbf{a}) - 0| \leq \eta \quad \vee \quad k = k_{\max} \quad (17)$$

where $k_{max}$ is the maximum number of iterations.

As a final comment, if $|g(\beta^k|\mathbf{a})| = 0$, for any $k \geq 0$, notice that according to (10) the increment is null and $\Delta_{\max}^k$ is unbounded, as expected, even if the GRM converges.

*D. Region Zooming Mechanism*

After being repaired, not all solutions will satisfy the equality constraint of the HL-RI problem. Those must be penalized. But, given the size of the uncertainty space, it is not practical to keep searching along directions not pointing towards the failure surface.

The Region Zooming Mechanism (RZM) acts indirectly on the binary encoded genes, relative to the elements in vector $\mathbf{a}$. The operator is complementary to the GRM and has a twofold purpose: 1) to identify the *target region* of the uncertainty space (where the MPP is expected to lie); and 2) to focus the search process on the target region.

In the scope of the HL-RI problem, the target region is defined as follows.

**Definition 3:** Let $B(\mathbf{y_{MPP}}; \epsilon)$ be an $\epsilon$-neighborood centered at $\mathbf{y_{MPP}}$. The set:

$$\mathcal{Z} = \{\mathbf{y} \in Y : \mathbf{y} \in B(\mathbf{y_{MPP}}; \epsilon) \land g(T^{-1}(\mathbf{y})) = 0\}$$

is called the *target region*.

In practice, in implicitly formulated problems, the true target region is unknown. The RZM divides the overall search process into three stages, during which the upper and lower bounds of each $a_i$, with $i = 1, \dots, N$, are updated, aiming to achieve better approximations of the true target region.

*1) First Evolution Stage*

On the first evolution stage, the search space comprises the region between two hyperspheres of radius $\beta_{\min}$ and $\beta_{\max}$. The initial search region $\mathcal{Z}^0 \subseteq \mathbb{R}^N$ is thus defined as follows.

**Definition 4:** Let $\beta_{\min}, \beta_{\max} \in \mathbb{R}_{\geq 0}$. The set:

$$\mathcal{Z}^0 = \{\mathbf{y} \in Y : \beta \in [\beta_{\min}, \beta_{\max}], \mathbf{a} \in [-1,1]^N\}$$

is called the *initial search region*.

The lower and upper bounds $\beta_{\min}$ and $\beta_{\max}$ are generic parameters of the algorithm that can be used, in practice, to accelerate the convergence of the GRM when some insight of the problem exists. Ultimately, if $\beta \in \mathbb{R}_{\geq 0}$, then $\mathcal{Z}^0 = \mathbb{R}^N$.

The goal of the first evolution stage is to search for solutions that are potentially on $\mathcal{Z}$. Identifying good solutions in the uncertainty space is a binary process: either they are on the failure surface, or they are not. Consider the following.

**Definition 5:** A solution $\mathbf{y} \in Y$ is said of *high probabilistic failure content*, if $\beta \in [\beta_{\min}, \beta_{\max}]$ and $g(T^{-1}(\mathbf{y})) = 0$.

The joint application of the GRM and the RZM allows an objective identification of solutions with high probabilistic failure content, in the sense that they are expected to be found in at most $k_{max}$ genetic repairs (see (17)), with $\bar{\beta} \leq \beta_{\max}$.

*2) Second Evolution Stage*

Let $Y_Z$ be the set of all solutions of high probabilistic failure content. The transition to the second stage is made on generation $\bar{t} = t_1$, when $\#(\mathbf{E}^{\bar{t}} \cap Y_Z) = \#\mathbf{E}^{\bar{t}}$. The search is then confined to a reduced region, using the information collected by the solutions in the elite of the population.

**Definition 6:** Let $a_{i\,\min}^{\bar{t}}$ and $a_{i\,\max}^{\bar{t}}$ be the minimum and maximum values of each $a_i$, for $i = 1, \dots, N$, existing among the elite solutions in $\mathbf{E}^{\bar{t}}$. The set:

$$\mathcal{Z}^{\bar{t}} = \{\mathbf{y} \in Y : \beta \in [\beta_{\min}, \beta_{\max}] \land \mathbf{a} \in [\mathbf{a}_{\min}^{\bar{t}}, \mathbf{a}_{\max}^{\bar{t}}]\}$$

is called a *reduced search region*.

Upon reduction, the mixed genotypes are (re)coded using the new side-constrains, increasing the resolution of the search process, for the same number of bits. Also, to prevent the hypothesis of dimensions being reduced to a single point, or becoming too narrow, a minimum diameter is imposed to each side-constraint:

$$a_{i\,\max}^{\bar{t}} - a_{i\,\min}^{\bar{t}} \geq \Delta a, \quad i = 1, \dots, N \quad (18)$$

The population is partially restarted at random, while keeping and recoding the existing elite genotypes, to promote a balance between *exploitation* and *exploration* on the newly defined reduced region.

The search region $\mathcal{Z}^{\bar{t}}$ is updated at every $t_Z$ generations, to gradually improve the search resolution, while keeping a relatively low number of bits. The successive reduction of the search region also allows to reallocate (translate) the search space, to conform to the condition in (18). That is, if $(a_{i\,\max}^{\bar{t}} - a_{i\,\min}^{\bar{t}}) < \Delta a$, then:

$$a_{i\,\max}^{\bar{t}} := \frac{a_{i\,\max}^{\bar{t}} + a_{i\,\min}^{\bar{t}} + \Delta a}{2}, \quad a_{i\,\min}^{\bar{t}} := \frac{a_{i\,\max}^{\bar{t}} + a_{i\,\min}^{\bar{t}} - \Delta a}{2} \quad (19)$$

for $i = 1, \dots, N$.

*3) Third Evolution Stage*

The recurrent reduction of the search region loses value as soon as the solutions in the population start concentrating in a single region of the search space. Thus, the zooming process is only repeated while a minimum level of diversity in the population is preserved (please, see [16-18]).

On the third evolution stage, there is refinement. The population is left to evolve in the last reduced search region, until a prescribed number of generations $\Delta t$ is complete, counting from generation $t_1$, at which the first reduced search region was defined.

## IV. BASIS EVOLUTIONARY OPERATORS

Structural reliability methods must be applicable to innumerous design solutions. A set of diverse evolutionary operators is introduced, combining high selective pressure (supported by a strong elitist strategy) with highly disruptive operators. The benefits of such a polarized approach are discussed by Eshelman [20].

The following operators are the core of the proposed EA and are referred as the *basis evolutionary operators* (BEOs). The goal is to build an EA capable of being accurate in a large range of problems with the same parameter settings, while working with very small populations and genotypes [16-18].

### A. Random Initialization

The same initialization process is applied at different stages of the evolution: each genotype is initialized as an array $(\beta_{\min}, b(\mathbf{a}))$, with $b(\mathbf{a})$ sampled at random from the uniform distribution. In the first generation, the entire population is initialized. At later stages, only selected genotypes are reinitialized.

### B. 'Elitist' Parent Selection

The parent selection mechanism begins by sorting $\mathbf{P}^t$, according to the fitness of its elements, dividing it into two groups: the elite $\mathbf{E}^t$ and the remaining population $\mathbf{P}^t \setminus \{\mathbf{E}^t\}$.

The mechanism selects $n_\mathbf{B}$ parent-lists that comprise the mating pool, at each generation, and is characterized by two independent fitness-proportional selection processes: one parent selected from $\mathbf{E}^t$, and another from $\mathbf{P}^t \setminus \{\mathbf{E}^t\}$.

This selection mechanism is twofold: 1) it increases selective pressure towards the best fitted genotypes of the population; 2) it increases the selection probability of the weakest genotypes of the population.

*C. Uniform Crossover*

A list $\mathbf{B}^t \in \mathbb{S}^*$ is a $n_{\mathbf{B}}$-tuple containing the offspring genotypes, at generation $t$. For each parent-list, only one offspring is generated and stored in the offspring population $\mathbf{B}^t$, hence generating a total of $n_{\mathbf{B}}$ genotypes.

The *uniform crossover* (UC) operator consists of selecting at random from which parent solution each gene comes from. The process is biased by the crossover ratio $r_{uc}$, representing the probability of selecting genes from the elite parent.

In UC, genetic features are not heritable unless common to both parents. As such, it is a highly disruptive recombination operator with the ability to create new genetic linkages, induced by the conflicting genes of the parents. The greater the genetic difference between two parents, the more explorative UC is. It is, thus, expected that UC progresses from a more explorative dynamic, in earlier stages of evolution, to a more exploitative one, in later stages.

*D. Elitist Survivor Selection*

After genetic recombination, the entire population is extended to a list $\mathbf{P}^t \cup \mathbf{B}^t$. The *survivor selection mechanism* is responsible for the composition of the population of the next generation. The overall selection process is performed by three elitist operators.

*1) Similariy Control*

In later generations, the fittest genotypes often recombine with near relatives and the explorative ability of UC degenerates. The *similarity control* (SC) operator promotes diversity, by impeding very similar genotypes to coexist.

**Definition 7:** Two genotypes $s_1, s_2 \in \mathbb{R}_{\geq 0} \times \mathbb{B}^l$ have an equal variable if $b(a_i)_1 = b(a_i)_2$, for $i = 1, \dots, N$. Two genotypes are *similar* if the number of equal variables is larger than some $\epsilon_{SC} \leq N$.

After fitness-based ranking of $\mathbf{P}^t \cup \mathbf{B}^t$, each genotype is compared against the weaker ones, one at a time. For each pair of similar genotypes, the one with less fitness is eliminated. After comparison, the size of $\mathbf{P}^t \cup \mathbf{B}^t$ is recovered with the random initialization of new genotypes.

*2) $(n + n_B)$ Replacement Mechanism*

To achieve a monotone evolution, the genotypes in $\mathbf{B}^t$ and $\mathbf{P}^t$ compete for survival: the intermediate population $\mathbf{P}^t \cup \mathbf{B}^t$ is ranked by fitness, after which only the top $n_{\mathbf{P}}$ genotypes are selected to survive.

*3) Implicit Mutation*

The SC operator alone doesn't guarantee genetic diversity throughout the evolution. Eventually, the evolutionary process manages to produce genotypes that are as similar as possible, and the operator is no longer efficient in injecting diversity.

After restoring the original size of the population, the $n_{\text{bot}} < n_{\mathbf{P}}$ genotypes with smaller fitness value are eliminated from the population and initialized at random.

Contrarily to SC, the main goal of *implicit mutation* is to inject raw diversity into the population, affecting the evolution in the long term. Indeed, its combination with the *elitist parent selection* promotes the recombination between the elite and the newly generated individuals, at all generations.

## V. LIMIT-BEHAVIOR ANALYSIS

The combination of the BEOs with the GRM and the RZM results in a hybrid EA, efficient with only a few bits per variable and very small populations. It is named the *Hybrid micro-Genetic Algorithm* (HmGA) [16-18].

The GRM is a deterministic operator that stops in finite time. A total repair, i.e., convergence to the failure surface with error $\eta$, implies null violation, contrarily to a partial repair. From (7), it follows that with suitable penalty parameters any *totally repaired* genotype has higher fitness value than any *partially repaired* genotype. Thus, we restrict the following analysis to the BEOs and RZM, assuming Theorems 1 and 2 hold.

The RZM divides the algorithmic structure of the HmGA in three stages. On the first stage, the algorithm searches for the failure surface, using the GRM as the main mechanism. On the second stage, the RZM promotes a "guided zooming" on the search space. On the third stage, there is refinement. See Algorithm 1.

The formal limit-behavior analysis of EAs with non-conventional algorithmic structures is important to demonstrate their ability to converge to the global optimum, when $t \to \infty$. Given a finite search space, here $\mathbb{S} = \mathbb{B}^l$, then convergence when $t \to \infty$ implies convergence in finite time [21].

We begin by analyzing the limit-behavior of the BEOs. Please, accept the notation $f(s)$ as the value of the objective function calculated for a given genotype.

**Definition 8:** The evolution is said *monotone* if, $\forall t \in \mathbb{N}$, $\max(f(s): \forall s \in \mathbf{P}^{t+1}) \geq \max(f(s): \forall s \in \mathbf{P}^t)$.

Based on a Markov Chains analysis of EAs, Eiben *et al.* [22] established the following conditions for the stochastic convergence of this class of algorithms.

**Proposition 2:** Let $\mathbf{P}^0 \in \mathbb{S}^*$ be an arbitrary initial population and $\mathbb{S}_{\text{opt}} = \{s \in \mathbb{S} : s \text{ is an optimum of } f\}$. If:

(A) the evolution is *monotone*;

(B) $\forall t \in \mathbb{N}$ and $\varepsilon_t \in [0,1]$, such that $\prod_{t=0}^{\infty} \varepsilon_t = 0$, the probability $\Pr(\mathbf{P}^{t+1} \cap \mathbb{S}_{\text{opt}} = \emptyset \mid \mathbf{P}^t) \leq \varepsilon_t$ holds;

then, the evolution reaches an optimum with probability 1.

We are now in conditions to demonstrate the convergence of the BEOs.

**Theorem 3:** The BEOs converge to $\mathbb{S}_{\text{opt}}$ with probability 1.

**Proof:** Let $\mathbf{M}^t \subseteq \mathbf{E}^t$ be the sub-list containing the genotypes with the maximum fitness value, among the elite genotypes of an *ordered* extended population $\mathbf{P}^t \cup \mathbf{B}^t$, at any $t \in \mathbb{N}$. From the $(n + n_{\mathbf{B}})$ replacement mechanism, it follows that:

$$\mathbf{P}^{t+1} = \mathbf{P}^t \cup \mathbf{B}^t \setminus \{\mathbf{B}^t\} \Rightarrow \Pr(\mathbf{M}^t \subset \mathbf{P}^{t+1}) = 1 \quad (20)$$

for any $t \in \mathbb{N}$. Furthermore, the SC and the *implicit mutation* operators always preserve at least the best solution. Thus, the evolution is monotone.

Regarding condition (B), note that *implicit mutation* is equivalent to gene-by-gene mutation with mutation rate $r_m = 0.5$. Thus, the probability of transitioning from an $s \in \mathbb{B}^l$ to any other $s_1 \in \mathbb{B}^l$ with $k$ different bits is:

$$p_{\text{mut}} = r_m^k (1 - r_m)^{l-k} > 0 \quad (21)$$

for all $0 \leq k \leq l$. Therefore, at any $t \in \mathbb{N}$, the probability of transitioning from any $\mathbf{P}^t$ to another $\mathbf{P}^{t+1}$ having at most $n_{\text{bot}}$ elements in $\mathbb{S}_{\text{opt}}$, through mutation, is bounded from below:

$$p \geq \prod_{j=1}^{n_{\text{bot}}} p_{\text{mut},j} > 0 \quad (22)$$

where $n_{\text{bot}}$ is the number of genotypes undergoing mutation. Thus, the probability of <u>not</u> transitioning to an optimal population, in two consecutive generations, is bounded from above:

$$\Pr(\mathbf{P}^{t+1} \cap \mathbb{S}_{\text{opt}} = \emptyset \mid \mathbf{P}^t) \leq 1 - p \quad (23)$$

Given that the BEOs form a set of homogeneous operators (i.e., whose parameters do not change in time), the transition probabilities are constant, at any $t \in \mathbb{N}$. Hence, for any initial population $\mathbf{P}^0 \in \mathbb{S}^*$, the probability of <u>not</u> transitioning to an optimal population, after $t$ generations, as $t \to \infty$, is bounded from above:

$$\lim_{t \to \infty} \Pr(\mathbf{P}^t \cap \mathbb{S}_{\text{opt}} = \emptyset \mid \mathbf{P}^0) \leq \prod_{t=1}^{\infty} \varepsilon_t = 0 \quad (24)$$

where $\varepsilon_t = 1 - p$. Therefore, the BEOs converge to $\mathbb{S}_{\text{opt}}$ with probability 1. ∎

The limit-behavior of the BEOs only guarantees convergence of the HmGA, on each of the search regions $\mathcal{Z}^0$ and $\mathcal{Z}^{\bar{t}}$, with $\bar{t} \in \mathbb{N}$. It doesn't necessarily imply convergence to the MPP. The reduction of the search region, imposed by the RZM, produces sequences $(\mathcal{Z}^{t_1 + k t_Z})_{0 \leq k \leq n}$ such that:

$$\mathcal{Z}^{t_1 + (k+1) t_Z} \cap \mathcal{Z}^{t_1 + k t_Z} = \emptyset, \quad d(\mathcal{Z}^{t_1 + (k+1) t_Z}) \leq d(\mathcal{Z}^{t_1 + k t_Z}) \quad (25)$$

and $\mathcal{Z}^{t_1} \subset \mathcal{Z}^0$, inducing *local search* in the sense that each reduced region focuses the search on a subregion of the previous one, with smaller diameter $d$. However, contrarily to the usual local search methods, the RZM establishes a sequence of search problems to be solved at each search region. Hence, conditions must be given for convergence to the MPP, when $t \to \infty$.

**Theorem 4:** The HmGA converges to $\mathbf{y}_{\text{MPP}}$ with probability 1, when $t \to \infty$, if and only if $\mathbf{y}_{\text{MPP}} \in \mathcal{Z}^{t_1 + n t_Z}$.

**Proof:** The proof is immediate. Assuming convergence with probability 1 to the MPP is true, then $\mathbf{y}_{\text{MPP}} \in \mathcal{Z}^{t_1 + n t_Z}$ must be true. Similarly, if $\mathbf{y}_{\text{MPP}} \in \mathcal{Z}^{t_1 + n t_Z}$ is true, then from Theorem 3 it follows that the HmGA will converge with probability 1 to $\mathbf{y}_{\text{MPP}}$. ∎

The previous result makes no assumptions on the reduced search regions, other than imposing that $\mathbf{y}_{\text{MPP}}$ belongs to the last search region, during the third evolution stage. In practice, it allows that the successive search regions translate to include $\mathbf{y}_{\text{MPP}}$, in case $\mathbf{y}_{\text{MPP}} \notin \mathcal{Z}^{t_1}$. However, this behavior is not desired, because no assumptions on the limit-state function $g$ are made, other than continuity, and the primal problem (3) is not necessarily convex. Hence, translation of the search regions only leads to global convergence if $\mathcal{Z}^{t_1}$ is contained in a close neighborhood of $\mathbf{y}_{\text{MPP}}$ where the problem is convex.

A more appealing condition for convergence is intuitively derived.

**Theorem 5:** If:

$$\mathcal{Z} \subset \cdots \subset \mathcal{Z}^{t_1 + k t_Z} \subset \cdots \subset \mathcal{Z}^0, \quad \forall k \leq n \quad (26)$$

then the HmGA converges to the $\mathbf{y}_{\text{MPP}}$ with probability 1, when $t \to \infty$.

**Proof:** The relation implies that the *interest region* belongs to all reduced search regions produced by the algorithm, which in turn implies that $\mathbf{y}_{\text{MPP}} \in \mathcal{Z}^{t_1 + k t_Z}, \forall k \leq n$. Hence, Theorem 4 is satisfied. ∎

The previous condition is sufficient, but not necessary. It is clear from Theorem 4 that convergence to the global optimum is always limited by the locality imposed by each reduction of the search region. In practice, such limitation is controlled with an appropriate tunning of the parameters of the algorithm. The experimental results show that the HmGA is quite robust without changing its parameters and is capable of accelerating the search (evolution) by keeping very low numbers of bits per variable, in highly dimensional and nonlinear structural reliability problems [16-18].

## VI. Conclusion

In probabilistic structural mechanics, the Hasofer-Lind reliability index (HL-RI) problem is defined as an equality constrained problem of searching for the minimum distance from a point to a surface. In previous works [16-18], the authors proposed the hybrid micro-Genetic Algorithm (HmGA) for the global solution of the HL-RI problem, as an EA that works on a mixed real-binary genotype space and comprises a set highly elitist and highly disruptive stochastic operators, called the Basis Evolutionary Operators (BEOs), hybridized with two novel deterministic operators called the Genetic Repair Mechanism (GRM) and the Region Zooming Mechanism (RZM).

The GRM is a repair operator used to enforce the feasibility of the solutions in the search space, or to decrease their degree of violation. The RZM divides the algorithmic structure of the HmGA in three stages. On the first stage, the algorithm searches for the failure surface, using the GRM as the main mechanism. On the second stage, the RZM promotes a "guided zooming" on the search space, around the most promising regions. On the third stage, there is refinement.

In this article, the limit-behavior of the algorithm is analyzed. The GRM is described as a local line search operator that generates a sequence of partial sums for each solution it is applied to. It is shown to be conditionally stable. Conditions for its convergence are given and it is shown that the absolute convergence of the sequence of partial sums implies the convergence to the feasible domain in finite iterations.

Based on a Markov Chain analysis, the BEOs are proven to converge to the global optimal of a finite search space with probability 1, as the time $t \to \infty$. From the convergence of the BEOs, conditions for the convergence of the HmGA are given, considering the algorithmic structure imposed by the RZM. The respective convergence theorems are given.

The proposed algorithm is expected to set the basis for further developments in the probabilistic reliability assessment of real-scenario engineering structures, where gradient-based algorithms are expected to diverge, and Monte Carlo Simulation is too expensive. As future research the inclusion of multiple failure criteria will be investigated.

---

**Algorithm 1:** Hybrid micro-Genetic Algorithm

*Initialization*
1: Set algorithm parameters
2: Set $a_i \in [-1,1]$, for $i = 1,..,N$
3: Set initial search region $\mathcal{Z}^0$
4: $t \leftarrow 1$
5: Initialize $\mathbf{P}^t$ at random.
6:    *Apply GRM to every $s \in \mathbf{P}^t$*

*First Evolution Stage*
7: **Do**
8:    $t \leftarrow t + 1$
9:    Apply BEOs to $\mathbf{P}^t$
10:    *Apply GRM to every newly generated $s \in \mathbf{P}^t$*
11: **until** $\#(\mathbf{E}^t \cap Y_\mathcal{Z}) = \#\mathbf{E}^t$

*Second Evolution Stage*
1: $t_1 \leftarrow t$
2: **Do**
3:    Set $a_i \in [a_{i_{\min}}, a_{i_{\max}}]$, for $i = 1,..,N$
4:    Set reduced search region $\mathcal{Z}^t$
5:    Recode genotypes in $\mathbf{E}^t$
6:    Randomly generate $\mathbf{P}^t \backslash \{\mathbf{E}^t\}$
7:    *Apply GRM to every newly generated $s \in \mathbf{P}^t$*
8:    $t_{\text{ref}} \leftarrow t$
9:    **Do**
10:      $t \leftarrow t + 1$
11:      Apply BEOs to $\mathbf{P}^t$
12:      *Apply GRM to every newly generated $s \in \mathbf{P}^t$*
13:    **until** $t - t_{\text{ref}} > t_\mathcal{Z}$
14:    $t \leftarrow t + 1$
15: **until** loss of diversity in $\mathbf{P}^t$ [16-18]

*Third Evolution Stage*
16: **Do**
17:    $t \leftarrow t + 1$
18:    Apply BEOs to $\mathbf{P}^t$
19:    *Apply GRM to every newly generated $s \in \mathbf{P}^t$*
20: **until** $t - t_1 > \Delta t$

21: **end**

ACKNOLEDGEMENTS

The authors gratefully acknowledge the funding by Fundação para a Ciência e Tecnologia (FCT), Portugal, through the funding of the "Associated Laboratory of Energy, Transports and Aeronautics (LAETA)".